# Development of a Forecasting and Warning System on the Ecological Life-Cycle of Sunn Pest

İsmail Balaban
*METU Department of Computer Engineering*
ismail.balaban@metu.edu.tr

Fatih Acun
*METU Department of Computer Engineering*
acun.fatih@metu.edu.tr

Onur Yigit Arpalı
*METU Department of Computer Engineering*
arpali.onur@metu.edu.tr

Furkan Murat
*METU Department of Computer Engineering*
murat.furkan@metu.edu.tr

Numan Ertuğrul Babaroğlu
*Directorate of Plant Protection Central Research Institute*
numanertugrul@tarim.gov.tr

Emre Akci
*Directorate of Plant Protection Central Research Institute*
emre.akci@tarim.gov.tr

Mehmet Çulcu
*Directorate of Plant Protection Central Research Institute*
mehmet.culcu@tarim.gov.tr

Mümtaz Özkan
*Directorate of Plant Protection Central Research Institute*
mumtaz.ozkan@tarim.gov.tr

Selim Temizer
*METU Department of Computer Engineering*
temizer@ceng.metu.edu.tr

*Abstract -* *We provide a machine learning solution that replaces the traditional methods for deciding the pesticide application time of Sunn Pest. We correlate climate data with phases of Sunn Pest in its life-cycle and decide whether the fields should be sprayed. Our solution includes two groups of prediction models. The first group contains decision trees that predict migration time of Sunn Pest from winter quarters to wheat fields. The second group contains random forest models that predict the nymphal stage percentages of Sunn Pest which is a criterion for pesticide application. We trained our models on four years of climate data which was collected from Kırşehir and Aksaray. The experiments show that our promised solution make correct predictions with high accuracies.*

INTRODUCTION

Sunn Pest is a widespread kind of wheat pest observed in the Middle Eastern Region. Generally, it appears in the various time slots of the year depending on the weather conditions. Sunn Pest gives substantial harm to wheat and significantly decreases the quality of bread to be produced with damaged wheat [1].

To get rid of Sunn Pests, fields are sprayed just before the insect causes a considerable loss on the wheat quality. However, determining the critical phase for this operation is not a trivial task since its arrival time to wheat fields differs in the year and the increase in its population is not a periodic event. Because the spraying time plays the most crucial role for the pesticide to be effective, it is essential to develop a solution on how to find the exact application time.

The current solution of detecting the pesticide application time of Sunn Pests is performed by physically controlling and counting Sunn Pests. The process of monitoring its life-cycle starts with the spring period and continues until the middle of the summer. Firstly, the time when Sunn Pests starts to travel from winter quarters to wheat fields is detected. Then, the Sunn Pests is counted by randomly putting a quarter square meter frame into wheat fields. During this counting process, agricultural engineers need to identify the 5 different nymphal stages of the insect [2]. Therefore, it is highly open to human errors.

In this paper, we present an autonomous solution that includes a software system and machine learning models depending on climate data to predict the phases of Sunn Pest in its life-cycle and the pesticide application time. Firstly, we collect the data from the METOS [3] weather stations deployed on the wheat fields and winter quarters. After doing preprocessing operations on this data, we train two groups of models. The first group consists of decision tree models that classify the phase of the Sunn Pest and the second group contains random forest models that predict nymphal stage ratios. Next, we fetch daily climate data and our models make predictions. Lastly, we display the results on sunetahminuyari.com to allow each farmer to reach these predictions.

The main contributions of this work are as follows:
- We provide machine learning models that determine the phases of Sunn Pest and pesticide application time.
- Our solution minimizes the economic loss due to human errors during the counting process.
- Our models are able to maintain and improve themselves within years without any human interaction.

TABLE 1 CLIMATE DATA

| Abbreviation | Description | Unit |
|---|---|---|
| WD | Wind Direction (avg) | ° |
| WS | Wind Speed (avg, max) | Km/h |
| SR | Solar Radiation (avg) | $W/m^2$ |
| R | Rainfall | Mm |
| D | Dewpoint (min, avg) | °C |
| RH | Relative Humidity (min, avg, max) | % |
| AT | Air Temperature (min, avg, max) | °C |

RELATED WORK

With the growing popularity of machine learning techniques, studies and technological practices made on agriculture has shown an increase. Our approach is differentiating from the proposed methods by preventing the Sunn Pests damage and reducing the cost of the prediction of pesticide application time operation.

Sunn Pests spread noise within the certain frequency range. Yazgaç et al. [4] can classify the Sunn Pests using these noise data, creating different feature vectors and applying different signal processing methods. With this technique, an embedded system was developed to detect Sunn Pests [5]. The number of Sunn Pests can be counted successfully with this method. However, this approach forces us to make measurements on the field and to predict the correct time to pesticide, we need to measure Sunn Pests regularly. So, it just proposes to replace human power using on the field to count Sunn Pests. Besides, Basati et al. which focuses on Sunn Pest-damaged wheat instead of directly Sunn Pests are creating models using pattern recognition methods [6]. In Turhal et al. research [7], instead of detecting the effects of Sunn Pest on wheat with eye observation, an automatic system which uses artificial neural networks is proposed to detect damaged wheat grain. Thus, the works that directly concern with Sunn Pest or its effect on wheat does not propose a method to solve the problem from beginning to end, neither the proposed methods avoid the consequences which Sunn Pest cause.

Differently, from the above studies on Sunn Pest and wheat, the researches indicating the usage of decision trees on agricultural and natural issues like our Sunn pest problem [8,9] assert that decision trees bring to a successful conclusion.

METHODOLOGY

In this section, we describe the data preprocessing steps and machine learning models that are used in phase classification and nymphal stage ratio regression.

Sunn Pest has three phases in its life-cycle [10]. The first phase corresponds to the period when it spends in winter quarters. The second phase is the migration period from winter quarters to wheat fields. The third phase describes the time interval when Sunn Pest is in wheat fields. We use a decision tree model to classify the phase of Sunn Pest in its life-cycle.

During the period in wheat fields, Sunn Pest has 5 different nymphal stages according to its matureness level [11]. The data provided by the Ministry of Agriculture and Forestry of Turkey is daily counts of these nymphal stages. In our project, we express these counts by percentages among nymphal stages as label data. We use a random forest model for regression to predict percentages of nymphal stages.

*I. Data Collection & Preprocessing*

We used climate data collected from weather stations in wheat fields and winter quarters. These stations are manufactured by METOS company, owned by the Ministry of Agriculture and Forestry. A subset of the dimensions of the climate data is described in Table 1.

Four-year climate data was collected from winter quarters and fields in Aksaray and Kırşehir. Also, software modules are currently fetching daily climate data. Using raw data in models may result in misleading predictions because the evolution of Sunn Pest reveals a continuum instead of a process that is affected by sudden changes in climate. Therefore, we accumulated raw data to be consistent with biological facts [12]. Accumulation of raw data starts concurrently with the yearly life-cycle of the Sunn Pest and refreshes at every life-cycle beginning. Another preprocessing operation is interpolation of missing climate data. Since specific sensors of weather stations may get broken due to various reasons, data from these sensors could not be acquired until it is fixed. To fulfill the missing parts, we perform linear interpolation.

*II. Decision Tree Models*

A decision tree splits the data depending on its discriminative features. Accumulated climate data fields that are acquired after preprocessing are very suitable to use while classifying the phases according to conditions. We used decision trees to take advantage of this discriminative property of our features. Figure 1 explains the three phases of Sunn Pest versus accumulated SR.

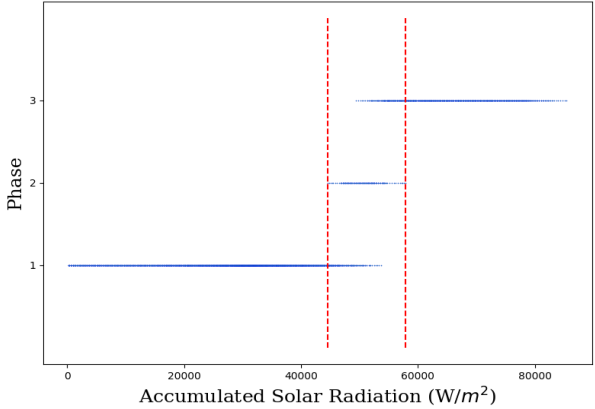

FIGURE 1 PHASE VERSUS ACCUMULATED SR

According to Figure 1, if accumulated SR is under 44533 W/m$^2$, the phase of insect is one. If its value is higher than 57912 W/m$^2$, the phase of insect is three. Otherwise, a decision tree can also classify the instances in between by looking at the other fields of the data which also expose similar characteristics as accumulated SR.

Another advantage of the decision tree is the model generated by it can be easily visualized. In our case, the visualized models were helpful to discuss the cases with agricultural engineers to verify the model whether it is consistent with the evolution of Sunn Pest.

### III. Random Forest Models

A random forest is constructed by a number of decision trees that vote for the output. We used the random forest for regression to predict continuous nymphal stage percentages. By making use of the voting mechanism between the trees in the forest, we reduce the possibility of over-fitting due to the relatively small size of our data. Furthermore, random forest is also suitable for our climate dataset due to the same reasons described in the previous part as it is based on decision trees.

## EXPERIMENTS

In this section, we present our experiments and their results. For predicting the phase of the Sunn pest, we used decision tree provided by the WEKA tool [13]. It uses C 4.5 algorithm [14] for constructing the decision tree. When building our model, we used decision tree with default parameters such that minimum object size in the leaf node is 1, minimum split size is 2, maximum depth is not limited, split quality measurement is GINI. For the regression model, we used random forest with default parameters provided by the WEKA tool such that number of trees in the forest is 10, split quality measurement is mean absolute error and rest of them is the same with decision tree.

Because our data set is small, we used 10-fold cross-validation to test our models. Test results provided in this document are based on this method.

TABLE 2 CONFIDENCE INTERVALS FOR 99%

|         | Decision Tree    | Random Forest    |
|---------|------------------|------------------|
| Model 1 | [0.1197, 0.1524] | [0.0273, 0.0343] |
| Model 2 | [0.0107, 0.0029] | [0.0020, 0.0029] |
| Model 3 | [0.0036, 0.0120] | [0.0028, 0.0039] |

*I. Phase Prediction*

Model 1 for phase prediction includes the features WS(avg, max), SR(avg), R, RH(avg), AT(min, avg, max), D(min, avg) of the climate data collected from wheat fields. This experiment has a low accuracy result, that is 86.3932% because it uses raw climate data. Thus, we do not use this model for real-time predictions.

Model 2 uses accumulated values of the fields that are used by Model 1. Therefore, it significantly raises the accuracy up to 99.3162% .

Model 3 narrows down the feature set used. We eliminated dewpoint and wind speed fields because they are not related to the evolution of Sunn Pest. Using a smaller subset of climate data also reduces the cost of weather stations because the number of sensors needed decreases.

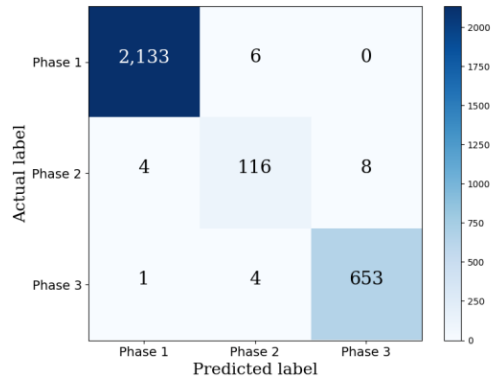

FIGURE 2 CONFUSION MATRIX OF MODEL 3

Figure 2 illustrates the classifier errors of Model 3. According to Figure 2, the number of incorrectly classified instances is 23 out of 2925, which means the model is reliable enough to make correct predictions with an accuracy of 99.2137%. Because the accuracy alone may be a misleading measure for the reliability, we calculate the confidence intervals according to Equation 1 to present a validation metric for population data. Table 2 indicates the confidence intervals of the models according to 99% confidence level.

$$e_p = e_s \pm z_N * \sqrt{\frac{e_s * (1 - e_s)}{n}} \quad (1)$$

*II. Pesticide Application Time Prediction*

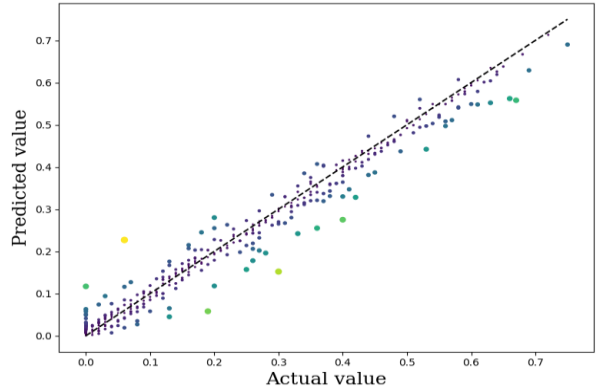

FIGURE 3 PREDICTED VERSUS ACTUAL VALUES OF NYMPHAL STAGE 2

For pesticide application time prediction, three experiments performed with the same fields of data as in phase prediction. The correlation coefficients of the three models are 0.6747, 0.9966, 0.9941 respectively. We also calculate the confidence intervals of random forest models by using Equation 2. Results are displayed in Table 2.

$$e_p = e_s \pm t_{n,d} * \frac{\sigma_s}{\sqrt{n}} \quad (2)$$

Figure 3 compares the predicted and actual values of $2^{nd}$ nymphal stage ratios. It can be observed that almost all of the instances placed very close to diagonal as expected.

## CONCLUSION

In this paper, we present a machine learning study on prediction of pesticide application time for the Sunn Pest. Our predictions are based on the climate dataset and Sunn Pest counting dataset which are given by Ministry of Agriculture and Forestry. We use a decision tree to classify the phases and random forest to determine the pesticide application time. According to the experiments, our models have accuracy over 99% and also offer credible confidence intervals. Unlike the traditional method that requires workforce that comes with its costs, our solution provides a much reliable process that can be easily maintained at a small expense. In addition, it can be extended to include other pests such as Aelia Rostrata, apple worm, and grasshopper through the cooperation of agricultural engineers.

In order to do exhaustive experiments for developing models further, the dataset should be enlarged. Therefore, having Sunn Pest data and climate data from a larger number of stations in Turkey will raise the project into a more accurate and robust state.